\documentclass[manuscript,nonacm]{acmart}

\usepackage{graphicx}
\usepackage{xcolor}
\usepackage{amsmath}
\usepackage{tabularx}

\AtBeginDocument{%
  \providecommand\BibTeX{{%
    \normalfont B\kern-0.5em{\scshape i\kern-0.25em b}\kern-0.8em\TeX}}}

\setcopyright{acmcopyright}
\copyrightyear{2018}
\acmYear{2018}
\acmDOI{10.1145/1122445.1122456}

\acmConference[PodRecs '20]{PodRecs: Workshop on Podcast Recommendations, 14th ACM Conference on Recommender Systems (RecSys)
}{Sept 25th, 2020} {Online (Previously Rio de Janeiro, Brazil)}

\begin{document}

\title{PodSumm: Podcast Audio Summarization}

\author{Aneesh Vartakavi}
\authornote{Both authors contributed equally to this research.}
\email{aneesh.vartakavi@nielsen.com}
\author{Amanmeet Garg}
\authornotemark[1]
\email{amanmeet.garg@nielsen.com}
\affiliation{%
  \institution{Gracenote Inc.}
  \streetaddress{2000 Powell Street, Suite 1500}
  \city{Emeryville}
  \state{California}
  \postcode{94608}
}

\renewcommand{\shortauthors}{vartakavi and garg}

\begin{abstract}
  
The diverse nature, scale, and specificity of podcasts present a unique challenge to content discovery systems. Listeners often rely on text descriptions of episodes provided by the podcast creators to discover new content. Some factors like the presentation style of the narrator and production quality are significant indicators of subjective user preference but are difficult to quantify and not reflected in the text descriptions provided by the podcast creators. 

We propose the automated creation of podcast audio summaries to aid in content discovery and help listeners to quickly preview podcast content before investing time in listening to an entire episode. In this paper, we present a method to automatically construct a podcast summary via guidance from the text-domain. Our method performs two key steps, namely, audio to text transcription and text summary generation. Motivated by a lack of datasets for this task, we curate an internal dataset, find an effective scheme for data augmentation, and design a protocol to gather summaries from annotators. We fine-tune a PreSumm\cite{liu2019text} model with our augmented dataset and perform an ablation study. Our method achieves ROUGE-F(1/2/L) scores of 0.63/0.53/0.63 on our dataset. We hope these results may inspire future research in this direction.


\end{abstract}


\maketitle

\section{Introduction}

The recent surge in popularity of podcasts presents a big opportunity and a unique set of challenges to existing content discovery and recommendation systems. Podcasts usually require active attention from a listener for extended periods unlike listening to music. Subjective attributes such as the speaker's presentation style, type of humor, or the production quality could influence the listener's preference but are hard to discern from a text description. 

In the video domain, movie trailers allow a viewer to preview some content and make a subjective decision to watch a film. The frequent release schedule for podcasts would make the production of such trailers for each episode impractical. Audio summaries have shown promise in improving the performance of spoken document search algorithms \cite{spina2017extracting}. We propose a method to create short podcast audio summaries in an automated manner. Such summaries could inform the listener about the topics of the podcast as well as subjective attributes like presentation style and production quality.


Podcasts present a unique set of challenges for an audio-based summarization algorithm. For example, podcasts usually focus on spoken word content and often contain overlapping speech from multiple speakers, free-form speech, audio effects, background music, and advertisements. A supervised learning algorithm operating in the audio domain would have to identify the nature of an audio segment before being able to judge its importance. This would require a large amount of training data manually annotated by listening to the audio in multiple passes, which is a difficult and time-consuming process.

However, as podcasts largely contain spoken-word content, summarization can also be performed in the text-domain on the transcript of an episode. In this work, we present our `Pod'cast `Summ'arization (\textit{PodSumm}) method to obtain podcast audio summaries guided by the text domain. PodSumm works by first transcribing the spoken content of a podcast, then, identifying important sentences in the transcript, and finally, stitching together the respective audio segments. We introduce a protocol and create an internal dataset specific to this task. In summary, we introduce the concept of podcast audio summaries to aid in content discovery. These summaries allow a listener to rapidly preview an episode before investing time in listening to the entire episode. 


\section{Related work} \label{section:related_work}

\subsection{Text summarization}
Neural models consider this task as a classification problem where a neural encoder creates a latent representation of the sentences, followed by a classifier scoring the sentences on their importance towards creating a summary \cite{nallapati2017, zhang2018neural}. With the rising popularity of Deep Neural Networks and Transformers \cite{vaswani2017attention}, pre-trained language models, particularly transformer models such as BERT \cite{devlin2019bert}, have shown promise in a wide range of NLP tasks. BERT can express the semantics of a document and obtain a sentence level representation. Recent approaches to text summarization like PreSumm \cite{liu2019text} and MatchSum \cite{zhong2020extractive} leverage BERT and achieve state-of-the-art performance on many benchmark datasets. These present a promising avenue for further development and expansion to other application domains.

\subsection{Speech summarization}
Speech summarization requires directly processing audio streams and providing snippets to create a combined audio summary. Prior solutions to this task have modelled this problem as a feature classification problem \cite{furui2003speech},a speech-text co-training problem \cite{xie2010semi} and graph clustering problem \cite{garg2009clusterrank}. Neural extractive summarization such as reinforcement learning \cite{wu2018learning}, hierarchical modeling \cite{liu2019hierarchical} and sequence-to-sequence modeling \cite{keneshloo2019deep} have shown promising results though on a limited variety of data. Automated speech summarization has many open research problems such as multi-party speech, spontaneous speech, handling disfluencies, and more.



\subsection{Podcast Summarization} 
There has been limited research on automated methods for podcast audio summarization. The diversity and narrative nature of podcasts with spontaneous speech, music, audio effects, and advertisements may present challenges for existing speech summarization methods. To address this issue, we pose the podcast audio summarization problem as multi-modal data summarization where we create an audio summary of a podcast with guidance from the text-domain.


\section{Our Method} \label{section:methods}

\subsection{PodSumm Architecture}

\begin{figure}[t]
    \centering
    \includegraphics[width=\textwidth]{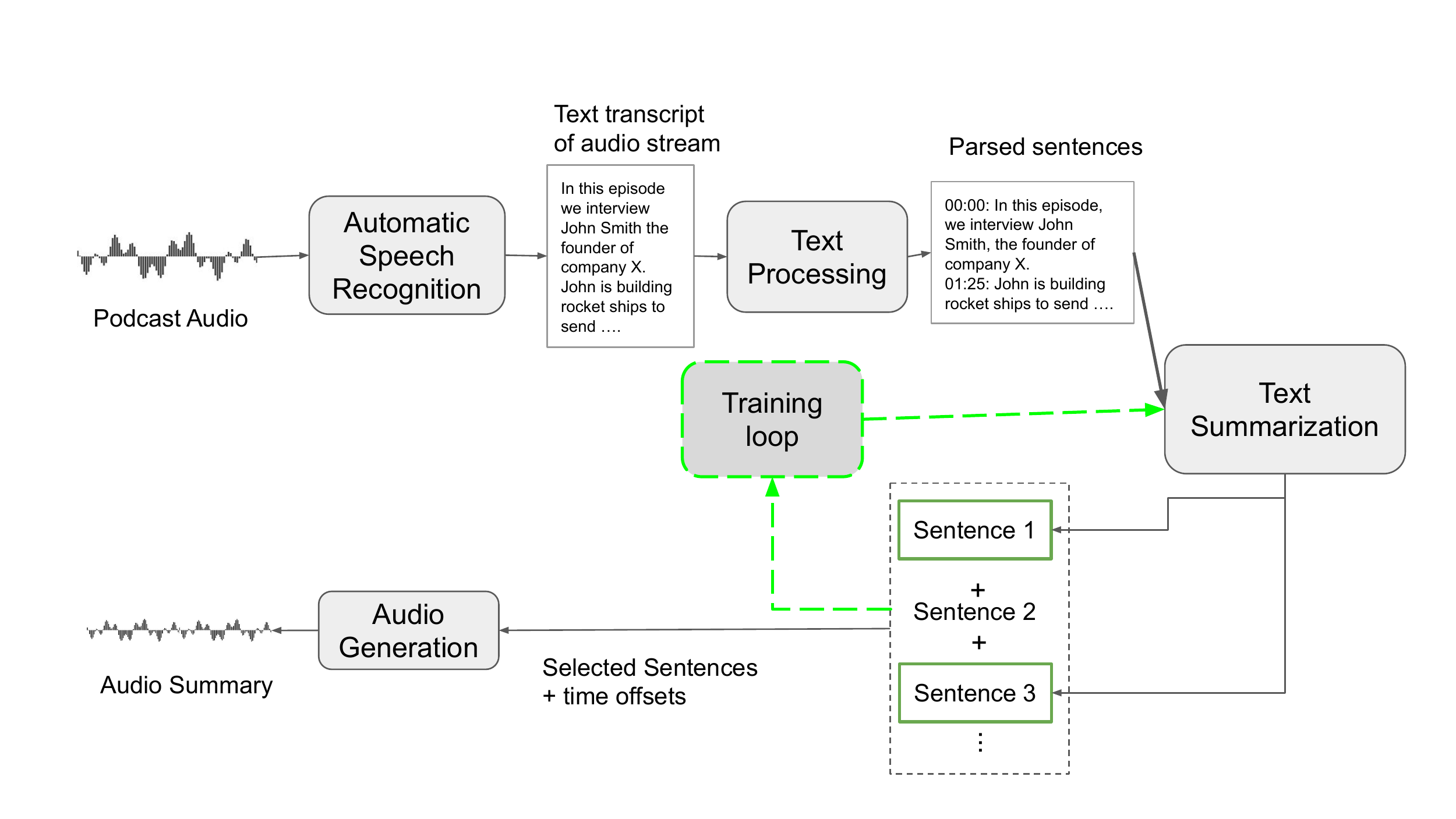}
    \caption{Block diagram of the PodSumm method and its modules.}
    \label{fig:PodSumm_blockdiag}
\end{figure}

The \textit{PodSumm} method comprises a sequence of steps, starting with the original audio stream and resulting in an audio summary obtained as an output (figure \ref{fig:PodSumm_blockdiag}). The first stage of the process is Automatic Speech Recognition (ASR), which generates a transcript. We then process the text to segment each podcast transcript into sentences. Subsequently, we use a fine-tuned text summarization model to select important sentences for inclusion in the final summary. We discuss each stage in detail below.

\subsubsection{Automatic Speech Recognition} \label{subsubsec:ASR}

ASR methods perform the task of speech-to-text transcription. They handle complexities related to varied accents, prosody and acoustic features, and speaker demographics. The quality of ASR transcriptions varies significantly and depends on the underlying training data \cite{Hannun2014DeepSS}. We leveraged a publicly available off-the-shelf solution (AWS Transcribe \footnote{https://aws.amazon.com/transcribe/}) \cite{di2019robust}, which allowed us to limit errors and focus on other core modules our pipeline.


\subsubsection{Text Processing}
The audio transcripts obtained in section \ref{subsubsec:ASR} above contain tuples of 1) text for individual words or punctuation marks 2) the start and end timestamps from the audio, and 3) a confidence score for the text prediction. We use Spacy \footnote{https://spacy.io/usage/linguistic-features\#sbd} to segment the text into sentences, and their corresponding start and end times in the audio. Additionally, we force a sentence break where a pause of over 2 seconds between words occurs. This helps us to better handle the cases where the ASR method missed a punctuation mark, which frequently occurs when music is played between speech segments.

\subsubsection{Text summary generation}
We then build text summaries by selecting appropriate sentences from the transcript, by leveraging advances in the field of extractive text summarization. We choose the PreSumm \cite{liu2019text} model, which builds upon BERT \cite{devlin2019bert} to obtain a sentence level encoding, and stacks inter-sentence transformer layers to capture document-level features for summarization.

We find that a PreSumm model pre-trained on the CNN/DailyMail dataset \cite{hermann2015teaching} does not produce adequate summaries for podcasts. Motivated by the lack of research datasets for this task, we created a dataset to further fine-tune the model for podcasts, described in Section \ref{section:dataset}.

The extractive PreSumm model performs summarization on a document with sentences $[sent_1 , sent_2, \dots, sent_m]$ by assigning a score $y_i \in [0,1]$ to each $sent_i$, indicating exclusion from or inclusion in the summary. The model is trained using a binary classification entropy loss to capture difference in prediction $\hat{y_i}$ and ground truth label $y_i$.

\subsubsection{Audio generation}
The predictions of the text summarization model include the sentence indices and respective scores. Using the stored sentence offsets, the audio representing the selected sentences are stitched together to obtain an audio summary.


\subsection{Dataset Creation} \label{section:dataset}
To address the lack of datasets for the task of podcast summarization, we curate a dataset to support the development and evaluation of our method. We selected 19 unique podcast series from different genres, selecting on average 16.3$\pm$6.28 episodes per series. The dataset contains a total of 188 hours of podcasts, with an average duration of 36.5 $\pm$19.8 minutes per episode. We built an annotation tool that presented the annotator with a sequence of sentences from the transcript of the episode, as well as the metadata from the podcast feed including the original audio of the episode. Each sentence was paired with the respective audio segment, derived using the offsets of each segment. Additionally, the annotation tool dynamically generated audio and text summaries based on annotator's selection enabling them to verify their choices.

The annotator was instructed to follow the protocol outlined below.
\begin{enumerate}
    \item \label{item:listen} Read the provider submitted description (if available) or listen to the audio of the podcast episode to understand the context and core message. 
    \item \label{item:select} Select the set of sentences that represent the summary of the podcast. The raters were requested to select continuous sequences of sentences where possible to minimize cuts in the audio while keeping the total summary length within 30-120 seconds.
    \item \label{item:playback}  Listen to the newly created sentence summary and repeat the above steps if necessary. Submit the annotations when a satisfactory summary is obtained.
\end{enumerate}

The resulting annotations include a set of sentence indices selected by the annotator as the most suitable candidates to create a summary. Due to resource limitations, each episode was annotated by a single annotator, due to which we are unable to compute the inter-annotator agreement. Discarding some outliers, we find that it took 3 minutes 57 seconds $\pm$ 4 minutes 51 seconds to annotate a single episode. We collected a total of 309 episodes with an average of 14.57 $\pm$ 7.01 selected sentences per summary.


\subsection{Model Training}

We begin with a PreSumm \cite{liu2019text} model pre-trained on the CNN/DailyMail dataset for 18000 steps \cite{hermann2015teaching} provided by the authors \footnote{https://github.com/nlpyang/PreSumm}, who report strong performance (ROUGE-$\mathit{(1,2,l)}=(43.85,20,34,39.90)$). We then fine-tune the model on our podcast dataset for 6000 steps as described in \cite{liu2019text}, beyond we noticed overfitting on our training set. The pre-trained model allows position embeddings of length 512, which we deemed sufficient for our application as the annotations in our dataset were contained within the first 512 tokens even for longer episodes.

Model checkpoints were saved and evaluated on the test set for every 1000 steps. The best performing model checkpoint was used for ablation experiments and to report system performance. For predicting summaries on new unseen data, we obtain the predicted scores for each sentence. Subsequently, top-$\mathit{n}$ sentences are selected from the rank-ordered candidates to create the final summary.


\subsection{Evaluation Metrics}
We report the precision, recall and F-measure for the ROUGE-1, ROUGE-2, ROUGE-$\mathit{l}$ scores \cite{lin-2004-rouge}. The metrics were selected to measure the ability of the model to produce summaries with overlapping words in comparison to the reference (recall), the prediction (precision), and average (F-measure). The $\mathit{n}$ in the ROUGE-$\mathit{n}$ metric signifies the unigram ($\mathit{n}$=1,single word overlap); bigram ($\mathit{n}$=2, consecutive word overlap) and longest common sequence overlap ($\mathit{n}$=$\mathit{l}$).

\subsection{Cross Validation Experiment} \label{section:cross_val}

Our current dataset consists of a total of 309 podcast episodes. This number is small in comparison to datasets such as CNN/DailyMail (312,102 data-label pairs) \footnote{https://github.com/abisee/cnn-dailymail} \cite{DBLP:journals/corr/SeeLM17, hermann2015teaching}.
To mitigate the effect of sampling bias, we report the mean and standard deviation of the ROUGE metrics from a $\mathit{k}$-fold ($\mathit{k}$=5) cross-validation experiment. The model was trained on the training split (80\% or 247 samples) and performance reported on the test split (20\% or 62 episodes), and the process was repeated for each fold.

\subsection{Data Augmentation}
We perform data augmentation to compensate for the relatively small size of our dataset and increase the generalization ability of the model. We observe that most previews and advertisements which should not be included in a summary are similar across podcasts episodes. We here describe a method to automatically find segments of repetitive content and our augmentation procedure.

We first find the indices of the sentences in the transcript that also occur in other episodes across our dataset (e.g.[0,1,2,3,4,6,7,30,31,32,48]). We then clean up the indices, 1) to merge any near-by indices ([0,1,2,3,4] with [6,7]) into one large set, and 2) to remove any outliers ([48]). All such repetitive content segments are stored for use in augmentations. To generate an augmented output, if an episode has repetitive content, we replace, else, we prepend the transcript with a randomly selected repetitive segment to create a new data sample. For each transcript, we add 20 new samples for the total augmented data size of 5166 samples for the training set for each fold.

\subsection{Ablation Studies}
\subsubsection{Effect of number of candidate sentences}
Similar to PreSumm, we select the top-$\mathit{n}$ sentences with the highest scores as our predictions. We study the effect of varying the number of sentences selected to represent the summary from the rank-ordered candidates in the model prediction. In our experiment, $\mathit{n} \in (5,9,12,15)$ was varied and the ROUGE-$\mathit{(1,2,l)}$ scores are reported.

\subsubsection{Effect of data augmentation}
The data augmentation applied during training alters the repetitive content preceding the sentences relevant to the summary. To test the effect of the data augmentation scheme on the model performance, we performed a fine-tuning experiment with and without data augmentation and report the system performance metrics.


\section{Results}

We summarize our results and ablation studies in Table \ref{table:results_fine_tuning}. As outlined in section \ref{section:cross_val}, we report the mean and standard deviation of the F-measure for the 3 metrics over the 5-fold cross validation experiment. Similar to prior work \cite{nallapati2017,zhong2020extractive,liu2019text}, we use a simple baseline "LEAD-\textit{n}", where we select the N leading sentences from the document as a summary.

 We find that LEAD-15 performs well, only slightly worse than a PreSumm model pre-trained on the CNN/DailyMail dataset with no-fine tuning. After fine-tuning on our dataset (PreSumm (FT ,$\mathit{k}=12$)), we find significant improvements in F-measure for all ROUGE-$\mathit{(1,2,l)}$ metrics over the baseline and the model with no fine-tuning. The model with augmentation, (PreSumm (FT + Aug ,$\mathit{k}=12$)) further improves performance, demonstrating that model performance on this task improves with even a small amount of task-specific data augmentation. 
 In the ablation study, we find that selecting the top ($\mathit{k}$=12) sentences produced the best results, compared to ($\mathit{k}$=5, 9 or 15).
 
\begin{figure}[ht]
    \centering
    \includegraphics[width=0.7\textwidth]{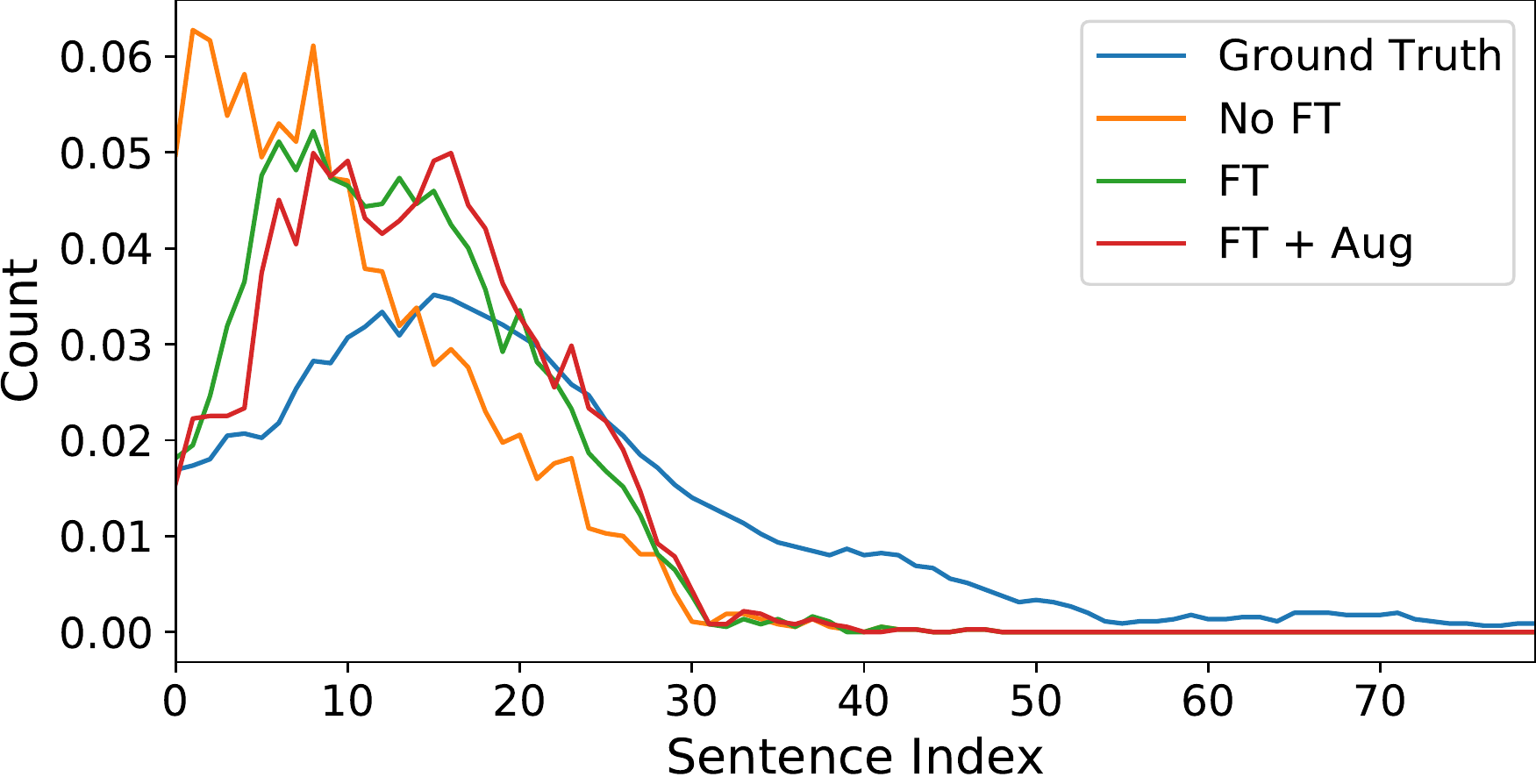}
    \caption{Selected sentence index vs. the normalized count (over all sentences in dataset) in the Ground Truth, predictions from PreSumm (No FT), fine-tuned PreSumm (FT), and fine-tuned PreSumm with augmentation(FT + Aug)}
    \label{fig:sentence_index_distance}
\end{figure}

 We display the distribution of sentence indices in figure \ref{fig:sentence_index_distance}. The ground truth data distribution indicates that the initial sentences with less related to the podcast summary task, which is corroborated by the relatively high performance of the LEAD-15 baseline relative to other LEAD-$\mathit{k}$($\mathit{k}$=5, 9 or 12) scores. We also see that the model without fine-tuning, PreSumm(No FT), is biased to select sentences from the beginning of the document, which is likely a property of the CNN/DailyMail dataset. The distributions after fine-tuning (FT, FT + AUG) are closer to the ground truth distributions, which are reflected in the metrics. However, the tails of these models still appear to follow the distribution model without fine-tuning. This highlights the need for further analysis and model development on a large dataset to account for all possible variations of the underlying data.
 
We present an example transcript along with the model predictions for PreSumm (no FT, $\mathit{k}=12$) in table \ref{table:example_summary_no_ft} and PreSumm (FT + Aug, $\mathit{k}=12$) in table \ref{table:example_summary_1}. The former model with no fine tuning selects a lot of sentences that are not relevant to the episode. In table \ref{table:example_summary_1} we see 9 true positive sentences (in green), 1 false-positive sentences (in blue) and 5 false-negative (in red) sentences, 11 repeated content sentences (in magenta), 2 of which were falsely predicted by the model (in cyan). This demonstrates that our method is able to correctly identify important sentences from the podcast transcription. The transcript also shows some errors that have accumulated through the system, eg. variations in spoken words (\textit{.org} mistranscribed as \textit{dot org's}), incorrect sentence segmentation between \textit{It is 7:30 p.m.} and \textit{On January,30th}, etc. Errors like these can complicate any downstream text processing, for example, a reader may only identify 3 false-positive sentences in the above example, whereas the system identified 5 due to incorrect sentence segmentation. 


\begin{table}[t]
\centering
\begin{tabular}{lccc}
\hline
Metric & ROUGE-$\mathit{1}$ & ROUGE-$\mathit{2}$ & ROUGE-$\mathit{l}$ \\ \hline

\multicolumn{4}{l}{Baseline} \\ \hline

LEAD-5 & 0.281 $\pm$ 0.019 & 0.166 $\pm$ 0.025 & 0.273 $\pm$ 0.018 \\
LEAD-9 &0.401 $\pm$ 0.029 & 0.257 $\pm$ 0.037 & 0.39 $\pm$ 0.028 \\
LEAD-12 & 0.465 $\pm$ 0.025 & 0.324 $\pm$ 0.032 & 0.455 $\pm$ 0.024 \\
LEAD-15 & \textbf{ 0.515 $\pm$ 0.026} & \textbf{ 0.389 $\pm$ 0.036} & \textbf{ 0.507 $\pm$ 0.026} \\ \hline 

\multicolumn{4}{l}{Fine-Tuning} \\ \hline
PreSumm (no FT, $\mathit{k}=12$) &  0.527 $\pm$ 0.016  & 0.381 $\pm$ 0.024 & 0.518 $\pm$ 0.017 \\

PreSumm (FT, $\mathit{k}=12$) & 0.625 $\pm$ 0.028 & 0.511 $\pm$ 0.034 & 0.619 $\pm$ 0.029 \\ 

PreSumm (FT + Aug, $\mathit{k}=12$) & \textbf{0.636 $\pm$ 0.022} & \textbf{0.529 $\pm$ 0.032} & \textbf{0.631 $\pm$ 0.023} \\
\hline

\multicolumn{4}{l}{Ablation - $\mathit{k}$ sentences} \\ \hline

PreSumm $\mathit{k}=5$ & 0.563 $\pm$ 0.027          & 0.458 $\pm$ 0.041          & 0.554 $\pm$ 0.029          \\
PreSumm $\mathit{k}=9$ & 0.626 $\pm$ 0.023          & 0.516 $\pm$ 0.033          & 0.62 $\pm$ 0.023           \\
PreSumm $\mathit{k}=12$& \textbf{0.636 $\pm$ 0.022} & \textbf{0.529 $\pm$ 0.032} & \textbf{0.631 $\pm$ 0.023} \\
PreSumm $\mathit{k}=15$ & 0.628 $\pm$ 0.018          & 0.528 $\pm$ 0.027          & 0.623 $\pm$ 0.019    \\ \hline


\end{tabular}
\caption{Results for the baseline, 5-fold cross validation experiment and 2 ablation experiments for the PreSumm method. The F-measure for ROUGE-$\mathit{(1,2,l)}$ metrics for pre-trained PreSumm model and the model fine-tuned with PodSumm dataset reported on the test set for each fold. Summary statistics for each metric reported as mean $\pm$ std. dev. over the 5 folds.}
\label{table:results_fine_tuning}
\end{table}

\begin{table}[ht]
\centering
\fcolorbox{black}{white}{
\begin{minipage}{\textwidth}
Hey there real quick before we start the show. California. \textcolor {blue}{We are coming your way for a live show next month on February 19th} \textcolor {cyan}{we are super excited to finally get to come to Southern California.} \textcolor {cyan}{We will be in 1000 Oaks with K C. L.} \textcolor {cyan}{You talking about the 2020 race and more to get your tickets, head over to NPR prisons dot org's.} \textcolor {cyan}{Oh, and if you're in Iowa, we have a show there tomorrow night Friday night, and there are still a few tickets available.} \textcolor {cyan}{Okay, here's the show. } \textcolor {cyan}{Hey there, it's the NPR politics podcast.} \textcolor {red}{It is 7:30} \textcolor {red}{p.m.} \textcolor {red}{On January 30th.} \textcolor {magenta}{I'm Tamara Keith.} \textcolor {cyan}{I cover the White House.} \textcolor {magenta}{I'm Aisha Roscoe.} \textcolor {magenta}{I also cover the White House} \textcolor {cyan}{and I'm Susan Davis.} \textcolor {magenta}{I cover Congress.} \textcolor {green}{Senate will convene as a court of impeachment today.} \textcolor {green}{The Senate impeachment trial is continuing with more questions and answers, senators asking questions, the House managers and the president's legal team answering those questions.} \textcolor {red}{And in fact, as we take this, the Q and A is still going on, so things could happen.} \textcolor {red}{That's why we do a time stamp.} \textcolor {red}{Um, Aisha, I'm wondering what stood out to you about today?} \textcolor {red}{Well, a lot of what the questions seem to be about was getting at this idea of.} \textcolor {green}{Is there a limit to what a president can do to get re elected?} \textcolor {red}{Because one of the president's lawyers representing him, Alan Dershowitz, made this argument that most presidents think their re election is in the public interest.} \textcolor {red}{And therefore, if they take actions to kind of help their reelection as long as it's not illegal, it's OK.} \textcolor {red}{And it really seemed like the senators were probing the limits of how far that argument can go on.}
\end{minipage}
}
\caption{PreSumm (no FT, $\mathit{k}=12$) output with correct predictions in \textcolor{green}{green}, false negatives  in \textcolor{red}{red} and false positive sentences in \textcolor{blue}{blue}. Sentences detected as repetitive content (\textcolor{magenta}{magenta}) and also falsely predicted by the model (\textcolor{cyan}{cyan})}.
\label{table:example_summary_no_ft}
\end{table}

\begin{table}[ht]
\centering
\fcolorbox{black}{white}{
\begin{minipage}{\textwidth}
Hey there real quick before we start the show. California. We are coming your way for a live show next month on February 19th \textcolor {magenta}{we are super excited to finally get to come to Southern California.} \textcolor {magenta}{We will be in 1000 Oaks with K C. L.} \textcolor {cyan}{You talking about the 2020 race and more to get your tickets, head over to NPR prisons dot org's.} \textcolor {cyan}{Oh, and if you're in Iowa, we have a show there tomorrow night Friday night, and there are still a few tickets available.} \textcolor {magenta}{Okay, here's the show. } \textcolor {magenta}{Hey there, it's the NPR politics podcast.} \textcolor {red}{It is 7:30} \textcolor {red}{p.m.} \textcolor {red}{On January 30th.} \textcolor {magenta}{I'm Tamara Keith.} \textcolor {magenta}{I cover the White House.} \textcolor {magenta}{I'm Aisha Roscoe.} \textcolor {magenta}{I also cover the White House} \textcolor {magenta}{and I'm Susan Davis.} \textcolor {magenta}{I cover Congress.} \textcolor {green}{Senate will convene as a court of impeachment today.} \textcolor {green}{The Senate impeachment trial is continuing with more questions and answers, senators asking questions, the House managers and the president's legal team answering those questions.} \textcolor {green}{And in fact, as we take this, the Q and A is still going on, so things could happen.} \textcolor {red}{That's why we do a time stamp.} \textcolor {red}{Um, Aisha, I'm wondering what stood out to you about today?} \textcolor {green}{Well, a lot of what the questions seem to be about was getting at this idea of.} \textcolor {green}{Is there a limit to what a president can do to get re elected?} \textcolor {green}{Because one of the president's lawyers representing him, Alan Dershowitz, made this argument that most presidents think their re election is in the public interest.} \textcolor {green}{And therefore, if they take actions to kind of help their reelection as long as it's not illegal, it's OK.} \textcolor {green}{And it really seemed like the senators were probing the limits of how far that argument can go on.} \textcolor {red}{At one point, there was a question from Senator Susan Collins from Maine, a Republican, and and a few other Republicans, including Senators Crepeau, Blunt and Rubio.} \textcolor {blue}{And remember, all of the questions are submitted in writing to the chief justice, who then reads them aloud.}
\end{minipage}
}
\caption{PreSumm (FT + Aug, $\mathit{k}=12$) output with correct predictions in \textcolor{green}{green}, false negatives  in \textcolor{red}{red} and false positive sentences in \textcolor{blue}{blue}. Sentences detected as repetitive content (\textcolor{magenta}{magenta}) and also falsely predicted by the model (\textcolor{cyan}{cyan})}.
\label{table:example_summary_1}
\end{table}


\section{Discussion}

In this work, we proposed PodSumm, a method to automatically generate audio summaries of podcasts via guidance from the text-domain. The method involves transcribing the audio, followed by some text processing and text summarization. An audio summary is then generated by stitching the audio segments that correspond to the sentences selected by the text summarization. The resulting model fine-tuned on our dataset performed better than a LEAD-N baseline and a model trained on the CNN/DailyMail dataset.


As our method contains a sequence of steps, the performance of each module directly influences the final produced audio summaries. In this paper we heavily leverage prior work in different fields, we believe custom modules would bring significant advantages. For example, a sentence segmentation model that is robust to transcription errors, or missing punctuation due to background music would allow us to leverage cheaper, less accurate ASR solutions. Further research is needed to develop and understand the effects of the individual modules specific to podcasts.

Although our proposed method showed improved performance after fine-tuning on our dataset, we recognize that its smaller size may restrict the generalization ability of the model on unseen data. Manual annotation of a large corpus of podcast data for this task is prohibitively expensive, but techniques like data augmentation could alleviate these to some extent.



\section{Conclusion}

We present a novel method to create audio summaries for podcasts via guidance from the text domain, and discuss the strengths and limitations. This work establishes the proof of working principle and sets direction for future development into a fully learned and automated method for podcast speech summarization. We look forward to newer methods emerging from the research community leading to an improved listener experience.


\begin{acks}
The authors thank Josh Morris for his counsel, Chinting Ko for his guidance on ASR, Joseph Renner, Jeff Scott, Gannon Gesiriech and Zafar Rafi for their feedback on the manuscript, and the contributions of the our team members at the Media Technology Lab at Gracenote.
\end{acks}

\bibliographystyle{ACM-Reference-Format}
\bibliography{references}

\end{document}